\pdfoutput=1

\documentclass[11pt]{article}

\usepackage[preprint]{acl}

\usepackage{times}
\usepackage{latexsym}

\usepackage{enumitem}

\usepackage{xcolor}

\usepackage[T1]{fontenc}

\usepackage[utf8]{inputenc}

\usepackage{microtype}

\usepackage{inconsolata}

\usepackage{graphicx}

\usepackage{amsmath}
\usepackage{amsfonts} 

\usepackage{tcolorbox} 
\usepackage{booktabs} 
\usepackage{subfigure}

\title{Beyond Self-Reports: Multi-Observer Agents for \\Personality Assessment in Large Language Models}

\author{Yin Jou Huang \and  Rafik Hadfi \\
  Graduate School of Informatics, Kyoto University, Kyoto, Japan\\
  \texttt{huang@nlp.ist.i.kyoto-u.ac.jp}, \texttt{rafik.hadfi@i.kyoto-u.ac.jp}\\}

\begin{document}
\maketitle
\begin{abstract}
Self‐report questionnaires have long been used to assess LLM personality traits, yet they fail to capture behavioral nuances due to biases and meta‐knowledge contamination. This paper proposes a novel multi‐observer framework for personality trait assessments in LLM agents that draws on informant‐report methods in psychology. Instead of relying on self-assessments, we employ multiple observer agents. Each observer is configured with a specific relational context (e.g., family member, friend, or coworker) and engages the subject LLM in dialogue before evaluating its behavior across the Big Five dimensions. We show that these observer‐report ratings align more closely with human judgments than traditional self‐reports and reveal systematic biases in LLM self-assessments. We also found that aggregating responses from 5 to 7 observers reduces systematic biases and achieves optimal reliability. Our results highlight the role of relationship context in perceiving personality and demonstrate that a multi-observer paradigm offers a more reliable, context-sensitive approach to evaluating LLM personality traits.
\end{abstract}


\section{Introduction}

\begin{figure*}[h]
    \centering
    \includegraphics[width=0.97\textwidth]{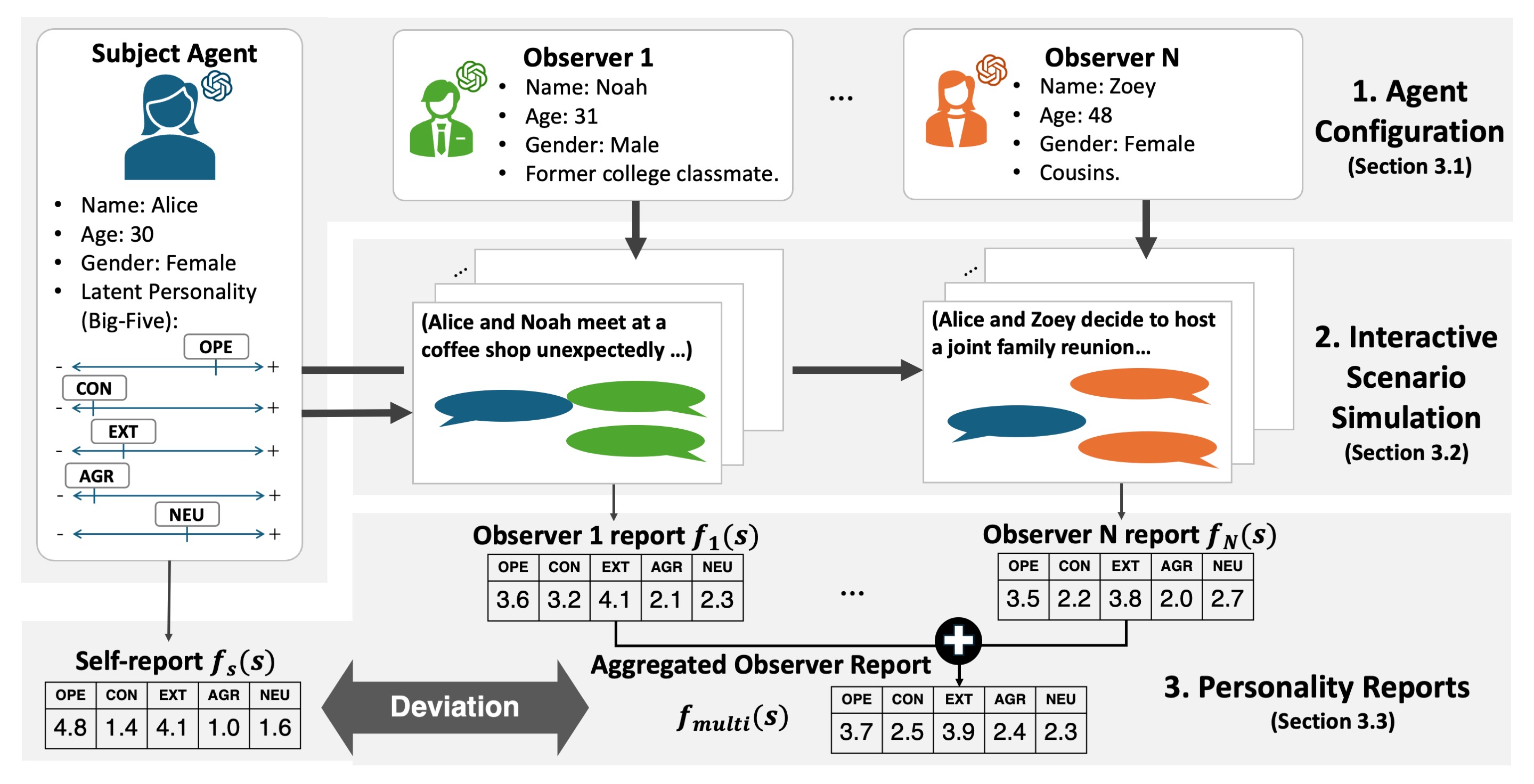}
    \vspace{-3mm}
    \caption{Overview: multi-observer LLM agents for Big Five personality assessment.}
    \vspace{-4mm}
    \label{fig:overview}
\end{figure*}

Large language models (LLMs) have demonstrated remarkable abilities to generate human-like responses and engage in complex social interactions~\cite{Kosinski2023, Lampinen2024}. 
Also, LLMs acquire the emergent ability of role-playing to emulate designated personas, leading to applications in fields like mental health support, education, etc~\cite{Lai2023, Hicke2023}.
As these LLM agents see wider deployment, there is a growing interest in assessing their personality traits~\cite{psychobench}. This task is crucial for a better understanding of their inherent characteristics and for developing more effective and appropriate human-AI interaction frameworks.

Human personality assessment has a long history in psychology, with various methods developed to evaluate individual traits and behaviors. Among these, self-report questionnaires are the most commonly used, assessing personality through individuals' responses to standardized questions about their thoughts, emotions, and behaviors. 
Existing LLM personality assessment methods also rely heavily on self-report questionnaires, in which an LLM is prompted to answer a set of standardized questions.
Despite their extensive use, researchers have raised questions about the reliability of using self-reports for LLM personality assessments, particularly in maintaining stable personality traits in different contexts~\cite{Gupta2023,dorner2023do,Wang2024c}. 
There are also potential risks of data contamination. It is likely that LLMs are exposed to discussions about personality tests during pretraining, but how this meta-knowledge influences the LLM’s test results remains unclear.

In this work, we explore alternative methodologies for LLM personality assessments.
We start from the intuition that others (such as friends, family, or colleagues) can provide valuable insights into a subject's personality traits, sometimes even more accurate than the subject themselves. 
In psychology, this intuition leads to the personality assessment method of informant-report~\cite{vazire2006informant}. Instead of using the self-report rating of the subject, informants other than the subject are asked to give ratings of the subject's personality.
In this fashion, it is possible to obtain a more objective assessment of the subject's personality profiles.


Inspired by the informant-report method, we propose a personality assessment framework that leverages multiple LLM agents as observers to report on the personality of a subject LLM agent (Figure~\ref{fig:overview}).\footnote{The code \& data will be made available upon publication.} 
For a given subject agent, we first prepare $N$ observer agents, each assigned a specific relationship with the subject, such as college classmates or cousins.
Based on this relationship, an array of interactive scenarios is generated automatically. 
For each scenario, we perform a simulation in which the subject and the observer agents engage in a dialogue based on the scenario.
After that, each observer agent is instructed to take a questionnaire and give ratings on the subject's personality from their perspective based on the dialogues obtained during the simulation process. 
Finally, all observer reports are aggregated to give a final collective assessment of the subject.

The experiments showcase that observer-report ratings align better with human ratings, while the self-report ratings reflect a high correlation with the injected personality prompt instead of real behaviors. Further, we empirically show that LLMs possess systematic biases in self-reporting their personality on some personality dimensions. 
We also analyze the influence of the number of observers ($N$) and the relationship between the observer and the subject. 
The analysis reveals the effectiveness of aggregating multiple observer responses to yield more robust personality ratings.



\section{Related Work}


The \textit{Big Five Inventory} (BFI) remains the most commonly used framework for evaluating personality traits, capturing the traits of openness, conscientiousness, extraversion, agreeableness, and neuroticism \cite{John1999}. Several variants of the BFI have also been utilized, including the BFI-2 \cite{Soto2017a}, BFI-2-XS \cite{Soto2017b}, and IPIP-NEO-120 \cite{johnson2014measuring}. Other prominent frameworks include the \textit{Myers-Briggs Type Indicator} (MBTI) \cite{Myers1962}, the \textit{HEXACO Personality Inventory} \cite{Lee2004}, and Goldberg’s bipolar adjective markers \cite{Goldberg1992}. Researchers have also investigated multi-rater assessments of personality traits. For instance, \cite{connelly2010other} showed how accurately various observers (e.g., friends, family members, coworkers) could rate an individual’s personality traits and how these ratings compare to self-assessments. Similarly, \cite{mount1994validity} explored the relationship between coworkers' personality ratings and job performance. Furthermore, \cite{vazire2010knows} investigated which traits are better judged by the self versus others, providing insights into the conditions under which external observations might outperform self-reports.


There is evidence for an alignment between self- and other-reports of personality. For instance, \cite{kim2019self} compared the Big Five self-ratings with informant ratings of the same individuals and found minimal differences in mean scores overall. Importantly, moderate discrepancies emerged only when the informants were strangers, implying that people tend to be more critical of individuals they do not know well. These results have important implications for personality assessment and contexts where self-enhancement motives may play a role. 

A comprehensive meta-analysis by \cite{connelly2010other} underscores the advantages of incorporating other-reports alongside self-reports for a richer and more predictive understanding of personality. Their findings indicate that accuracy in other-reports varies by trait. That is, extraversion and conscientiousness are rated most accurately, while emotional stability (neuroticism) and agreeableness are more difficult to identify, particularly for observers who lack familiarity with the target. One explanation for lower accuracy in agreeableness lies in its high evaluativeness, which can make ratings more subjective. Moreover, family and friends provide the most accurate other-ratings, while coworkers, despite frequent interactions, tend to offer less accurate assessments.

Regarding predictive validity, \cite{connelly2010other} also found that other-ratings often outperform self-ratings in predicting academic and job performance; multiple other-raters further enhance predictive power relative to a single informant. In contrast, self- and other-ratings exhibit comparable validity for forecasting first impressions. This led to the conclusion that there is ``extraordinary value'' in collecting other ratings of personality while emphasizing the importance of using multiple raters to mitigate individual bias, the importance of well-acquainted observers, the subtlety in evaluating less visible traits, and the importance of specific context and purpose of the assessment. Building on human personality trait assessments, we examine how these approaches translate into evaluating personality traits in LLMs \cite{safdari2023personality,huang2024reliability}.

Early personality assessments relied on predefined templates, which ultimately progressed to end-to-end dialogue models that encode fundamental persona traits \cite{Zhang2018}. Despite these advances, recent findings indicate that LLMs often fail to exhibit consistent personality scores when evaluated through standard self-report measures \cite{Gupta2023,tommaso2024llms}. \cite{Gupta2023} systematically confirmed the unreliability of human-oriented self-assessment methods applied to LLMs. 
Complementing these findings, \cite{Zou2024} investigated the misalignment between chatbot self-reports and user perceptions, asking whether LLM-based chatbots truly have valid, self-reported personalities. 
The results showed weak correlations between self-reports, user perceptions, and interaction quality, raising concerns about the predictive validity of LLM self-reports. 

\vspace{-1mm}
\section{Methodology}
\label{sec:methodology}

We now introduce our multi-observer framework of personality assessment. We first introduce the configuration of the subject and observer agents in Section~\ref{sec:agent}. We then utilize these agents to simulate scenarios (Section~\ref{sec:simulation}) before observers give the personality assessment on the subject (Section~\ref{sec:observer-report}).

\subsection{Agent Configuration} \label{sec:agent}
Our proposed personality assessment framework involves a subject agent and multiple observer agents. We introduce the configuration of the agents.

\paragraph{Agent Profile}
Each agent is given a randomly generated basic profile that contains \textbf{Name}, \textbf{Age}, and \textbf{Gender}.
Setting specific agent names facilitates smooth interaction in the simulation phase, and incorporating random age and gender of the agents results in a greater diversity of observers. 
For both subject and object agents, a randomly generated basic profile is assigned. 

For each subject agent $s$, we also assign an additional latent personality profile $\psi_s$.
In this work, we adopt the Big Five personality theory~\cite{Goldberg1992, John1999}, which decomposes human personality into five dimensions: openness (\textbf{OPE}), conscientiousness (\textbf{CON}), extraversion (\textbf{EXT}), agreeableness (\textbf{AGR}), and neuroticism (\textbf{NEU}). Based on Big Five, we define $\psi_s$ as a $5$-dimensional vector $(\psi_s^{OPE}, \psi_s^{CON}, \psi_s^{EXT}, \psi_s^{AGR}, \psi_s^{NEU})$. Each dimension $\psi_s^d$ of $\psi_s$ is an integer within the range of $[1, 6]$, indicating the strength level of a corresponding Big Five personality dimension $d$.

Following previous work, we construct a personality instruction of $\psi$ based on personality markers~\cite{Serapio2023}. For each dimension $d$, we pick $m$ personality markers that reflect the personality strength $\psi_s^d$. For instance, if the agent $s$ has an extraversion trait of strength $\psi^{EXT}_s = 2$, which is on the lower side of the spectrum, some possible choices of personality markers are ``timid,  silent, unsociable''.\footnote{See Appendix~\ref{appendix:agent_profile} for the details of the agent profiles.}


\paragraph{Relationship Generation} For each pair of subject and observer agents, we generate a relationship that matches their profiles. In previous psychology works, informant reports are often conducted by individuals who have a close relationship with the subject. Here, we follow ~\citet{kim2019self} and generate relationships within one of the following \textbf{relation contexts}: Family, Friend, or Workplace. Based on the agent profiles and a designated relation type, an inter-agent relationship that matches the profiles is generated. Here, we utilize a separate LLM to generate the relationships automatically. 

\subsection{Interactive Scenario Simulation} ~\label{sec:simulation}
Personality manifests through behaviors. How individuals react in different situations reflects their underlying personality characteristics. 
Unlike self-report questionnaires, which can be influenced by bias or social desirability, observing behavior in diverse scenarios allows for a more accurate and objective assessment.
Based on this motivation, we conduct simulations of a diverse set of scenarios involving the subject and the object agent to elicit their different behavioral patterns.
Specifically, a set of interactive scenarios is generated based on the profiles and the relationship between the subject and observer agents. The agents then engage in a dialogue based on these scenarios. In the following, we summarize the process.



\paragraph{Scenario Generation}

We next generate a set of $K$ scenarios involving the subject and observer based on their relationship. Specifically, we generate diverse scenarios that can elicit behaviors of the subject agent that signals various aspects of its personality.
Similar to the process for generating inter-agent relationships, we use a separate LLM to generate the scenarios that fit the above criteria. 

\paragraph{Interactive Scenario Simulation}
Further, we perform a simulation in which the subject and the observer agents engage in a dialogue based on the generated scenarios. For each scenario, we first configure the subject agent and the observer agent based on their profiles (Section~\ref{sec:agent}). Further, the agents are instructed to converse with each other based on the relationship and scenario settings. The agents generate utterances alternatively, with the observer always kick-starting the dialogue. Each generated utterance is fed to the other agent as a prompt to generate the next utterance. In addition to the utterance, the agents are asked to specify whether the dialogue is over or if they wish to leave the conversation. The simulation is terminated if both agents reply that the dialogue should be over. 


\subsection{Personality Reports}
\label{sec:observer-report}

We utilize a questionnaire to assess the personality of a specific subject agent.
The personality assessments are made from three perspectives: the subject's self-report, the individual observer-report, and the aggregated observer-report from a group of observers. The report of agent $n$ on agent $s$'s personality trait is represented as $f_n(s)$, a 5-dimensional vector representing Big Five personality traits \cite{huang2024rafik}.


\paragraph{Subject's Self-Report} We obtain the subject's personality assessment of itself.
Following previous studies, we use a personality test questionnaire containing $M$ statements such as ``being the life of the party'', ``sympathize with others' feelings''. For each statement, the subject agent $s$ is instructed to rate how accurately the statement describes itself using a 5-Likert scale (from ``1 = very inaccurate'' to ``5 = very accurate'').
See Appendix~\ref{appendix:prompt} for details.
In the questionnaire, each statement is associated with one of the Big Five personality dimensions. To assess the strength of dimension $d$, we calculate the average rating score of its related statements as the final assessment score $f_s^{d}(s)$, which is the $d$th dimension of the subject's self-report $f_s(s)$.


\paragraph{Individual Observer-Report} We obtain the subject's personality assessment from the perspective of each observer agent $n$. Similar to the subject's self-report, we instruct the observer to rate each statement in the questionnaire on a scale of 1 to 5. In the case of observer reports, the observer agent is asked to rate how each statement fits the description of the subject agent.
To get the assessment from the perspective of the observer agent, the dialogues generated from the scenario simulation phase are also provided in the prompt. 
In this fashion, we obtain the individual observer-reports $f_n(s)$ of each observer agent $n$ on agent $s$.

\paragraph{Aggregated Observer-report} Given a group of $N$ observers, we calculate the aggregated multi-observer report based on the individual ratings. In this work, we simply take the average value of the observer reports as $f_{multi}(s) = \frac{1}{N} \sum_{n=1}^{N} f_n(s)$. This aggregated observer report reflects the collective reports of all observers \cite{fleenor2006wisdom,burton2024large}. Since each observer’s evaluation is inherently subjective and shaped by their unique relationship with the subject, we expect that combining these perspectives will yield a more reliable measure of the subject’s personality by reducing individual biases of single agents.





\section{Experimental Settings} \label{sec:exp}
We provide details on the experimental settings.

\vspace{-2mm}
\paragraph{LLM Agents} 
For the subject and observer agents, we adopt GPT-4o as LLM~\cite{hurst2024gpt}. We also conducted experiments based on Qwen2.5 and Llama-3, found in Appendix~\ref{appendix:llama}. 

\vspace{-1mm}
\paragraph{Scenario Generation} We use the GPT-4o model to generate inter-agent relationships and scenarios. We perform scenario simulation based on a total of $100$ subject agents. For each subject agent, we assign $N=15$ observer agents.
Among the $15$ observers, $5$ have friend relationships, $5$ have family relationships, and $5$ have relationships within the workplace context. For each pair of subject and observer agents, we generate $K=5$ scenarios and conduct the simulation.\footnote{Refer to Appendix~\ref{appendix:prompt} for the details of the prompts.}

\vspace{-1mm}
\paragraph{Personality Questionnaire} We adopt the International Personality Item Pool (IPIP) personality test~\cite{goldberg1992development}, which is a widely used personality inventory designed for assessing the Big Five personality traits. The questionnaire consists of $M=50$ statements, with each statement related to one of the Big Five personality dimensions.\footnote{
The list of 50 items and the scoring schemes can be found at \url{https://ipip.ori.org/newBigFive5broadKey.htm}.}




\section{Results} \label{sec:results}
\vspace{-2mm}
In this section, we present the experimental results of our personality assessment method.

\subsection{Validity of Observer-Report} \label{sec:basic_results}
\vspace{-1mm}

We start by verifying the validity of our proposed personality assessment method via observer reports.
\begin{figure}[t]
    \centering
    \includegraphics[width=0.99\columnwidth]{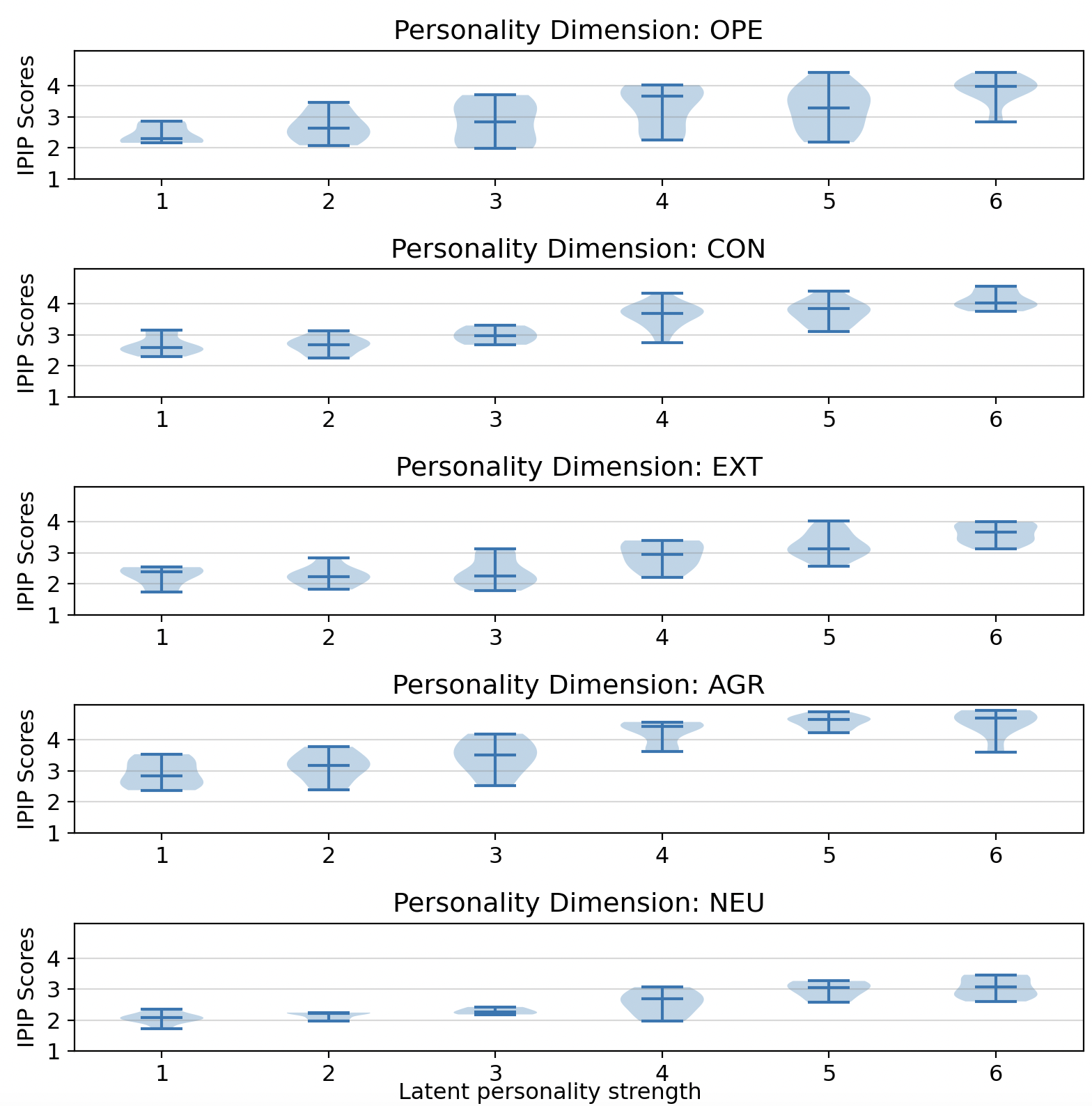}
    \vspace{-6mm}
    \caption{Observer report ratings across different latent personality strength levels.} 
    \vspace{-1mm}
    \label{fig:observer_levels}
\end{figure}
Figure~\ref{fig:observer_levels} shows the change in observer-report scores across agents with different levels of latent personality strengths ($\psi^d_s$) in each Big Five dimension. For all dimensions, we observe an increasing tendency of observer rating scores (\textbf{IPIP Scores}) with the increase of latent personality strength level.

\begin{table}[t]
\centering
\resizebox{\columnwidth}{!}{%
    \begin{tabular}{lccccc}
    \toprule
     & \textbf{OPE} & \textbf{CON} & \textbf{EXT} & \textbf{AGR} & \textbf{NEU} \\
    \midrule
    latent-self         & 0.97 & 0.95 & 0.95 & 0.94 & 0.93 \\
    latent-observer     & 0.55 & 0.85 & 0.84 & 0.84 & 0.86 \\
    \midrule
    human-self         & -0.25 & \textbf{0.47} & \textbf{0.79} & 0.63 & 0.22 \\
    human-observer     & \textbf{0.48} & 0.43 & 0.76 & \textbf{0.85} & \textbf{0.42} \\
\bottomrule
\end{tabular}
}
\vspace{-3mm}
\caption{Spearman’s rank correlations}
\vspace{-4mm}
\label{table:corr_compare}
\end{table}

\vspace{-2mm}
\paragraph{Correlation with Latent Personality Profiles}
We compare the self- and observer-report ratings in terms of their correlation with the latent personality strengths in each Big Five dimension. We precisely define the \textbf{latent-observer} case as correlating the subject's latent personality $\psi_{s}$ with the aggregated observer report $f_{\text{multi}}(s)$. On the other hand, the \textbf{self-observer} case correlates the subject's self-report $f_{\text{s}}(s)$ with the aggregated observer report $f_{\text{multi}}(s)$. The first row of Table~\ref{table:corr_compare} shows the resulting correlation values. For all personality dimensions, the self-report ratings show a correlation coefficient exceeding 0.9, showing a near-perfect positive correlation with the latent profiles. On the other hand, the correlation strength between observer-report ratings and the latent personality strength levels is lower.

\paragraph{Correlation with Human Ratings}
We crowd-sourced a small-scale human rating dataset in which human annotators are asked to provide their personality ratings of a subject agent based on the dialogues between the subject and observer agents.\footnote{See Appendix~\ref{sec:human} for the data collection process.}
We then calculated the agreement between the human and self- and observer-report ratings, respectively. The lower section of Table~\ref{table:corr_compare} shows the Spearman's rank correlation coefficients. For the openness, agreeableness, and neuroticism dimensions, the observer-report ratings correlate more with human ratings. While self-reports correlate more with human ratings for conscientiousness and extraversion, the difference is marginal. Notably, self-report exhibits a negative correlation for openness, implying a reversal in rank ordering relative to human ratings. With these results, we verify the validity of the proposed observer-report personality assessment method, which aligns better with human ratings than the self-report method. The results also raise concerns about the reliability of the self-report method. The relatively lower correlation with human ratings indicates that the near-perfect correlation with the latent personality profile only reflects the personality instruction prompts, but not the actual behavior of the agent.

\subsection{Impact of Multiple Observers} \label{sec:result_n_observer}

\begin{figure*}[h]
    \centering
    \includegraphics[width=1.02\textwidth]{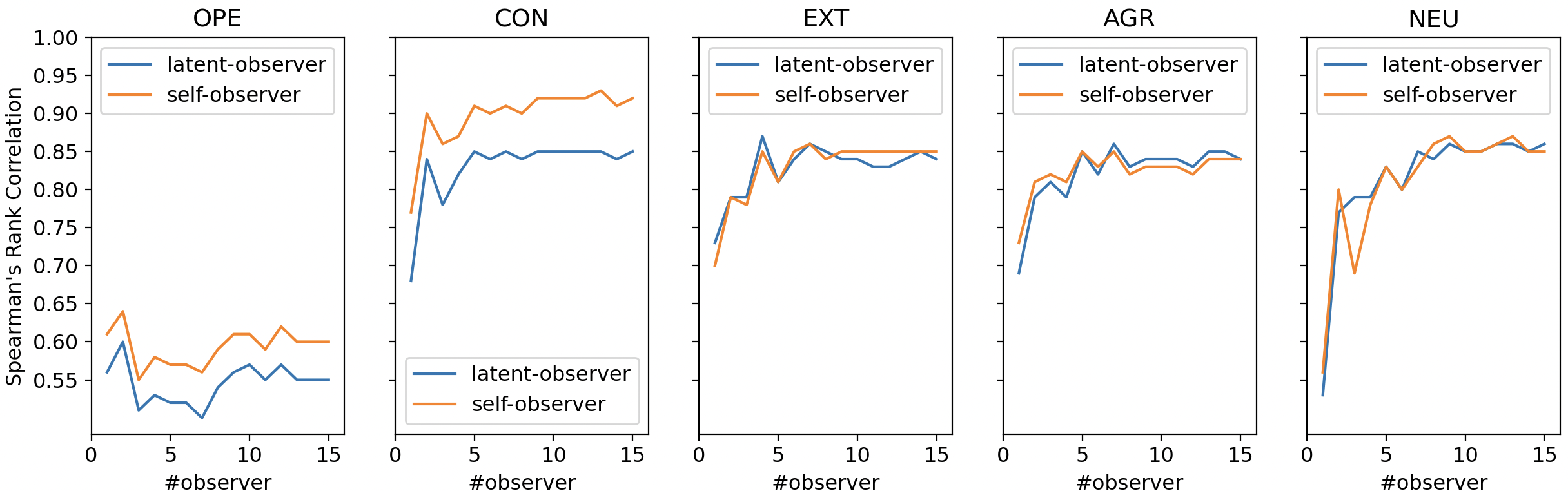}
    \vspace{-5mm}
    \caption{Spearman's Rank Correlation coefficients between latent-observer and self-observer ratings as a function of the number of observers for each Big Five trait.}
    \vspace{-2mm}
    \label{fig:fig2}
\end{figure*}

We now examine how the number of observers influences the personality assessment outcomes. We compute Spearman's rank correlation coefficients for the latent-observer and self-observer cases. Figure \ref{fig:fig2} illustrates how these correlation values vary with the number of observers, ranging from 1 to 15, for each Big Five dimension. Conscientiousness improved with additional observers, with both correlation types increasing sharply to five observers. Self-observer correlations stabilized at 0.90, while latent-observer correlations stabilized at 0.85, making it the personality trait with the highest agreement. Extraversion and agreeableness demonstrated an increased correlation pattern with additional observers before both correlation values converged at 0.85. Neuroticism displayed the most interesting convergence pattern, initially variable, with correlations reaching similar levels of 0.85 for both correlation types when incorporating around seven observers. Openness consistently showed the lowest correlations among all traits (0.60 for latent-observer, 0.65 for self-observer), with minimal improvement from additional observers. This suggests that openness may be more challenging to rate, regardless of the number of observers.


\paragraph{Trait Visibility} 


Many factors affect the agreement between self- and observer-ratings of personality traits. A key factor is trait visibility, which is the extent to which a trait is expressed through overt behavior that can be easily observed by others~\cite{funder1995accuracy}. Traits with high visibility, such as extraversion, tend to show stronger self–other agreement. Conscientiousness in particular yields the highest self–observer agreement~\cite{connelly2010other}. 
In contrast, openness is associated with internal characteristics such as imagination, aesthetic sensitivity, curiosity, etc. Due to the low trait visibility, openness often demonstrates the lowest agreement \cite{vazire2010knows}. Our results align with these findings, with the high agreement for conscientiousness and low agreement for openness. However, for neuroticism, another trait with lower trait visibility, we do not identify a lower self-observer agreement.


\paragraph{Wisdom of the Crowd}
Another finding is that the benefits of additional observers generally stabilize after 5-7 observers. 
Below 5-7 observers, we observe something akin to the wisdom of the crowd effect, where groups outperform top individuals by pooling diverse insights and aggregating responses from multiple sources to generate superior outcomes compared to relying on a single model~\cite{burton2024large,guo2024large,fleenor2006wisdom}.
This insight sets our multi-observer framework for more robust personality assessments.
Beyond 5-7 observers, additional observers introduce diminishing returns for capturing an LLM's personality consistently, suggesting this as an optimal number for practical assessment purposes.
This threshold recalls Dunbar's number, where human social networks are naturally organized in layered structures. In particular, the innermost layer, often referred to as the support clique, consists of five individuals on average \cite{dunbar2015structure,hill2003social,roberts2011costs}. This observation aligns with research showing that intimate bonds (family, close friends, or trusted colleagues) offer the most revealing insights into an individual's personality. Real-world social structures demonstrate that a core group of just a few relationships is sufficient to capture the most profound knowledge of a person's traits.

\subsection{Self-Observer Deviations} \label{sec:deviation}

We calculate the differences between aggregated self- and observer-report scores. We also identify systematic biases between self- and observer-report ratings via mean deviation $\frac{1}{N} \sum_{s=1}^{N} f_{multi}(s) - f_{s}(s)$.
Positive values indicate higher observer ratings than self-ratings, and negative values indicate lower observer ratings. Zero or near‐zero values imply close agreement.
\begin{table}[t]
    \centering
    \resizebox{\columnwidth}{!}{%
    \begin{tabular}{lccccc}
    \toprule
      & OPE & CON & EXT & AGR & NEU \\
    \midrule
    Mean Deviation & 0.20 & \textbf{0.39*} & -0.12 & \textbf{0.91*} & -0.19 \\
    Cohen’s d (LLM) & 0.24 & 0.46 & -0.13 & 1.07 & -0.26 \\
    \midrule
    Cohen’s d (human) & 0.27 & 0.27 & 0.21 & 0.26 & 0.13 \\
    \bottomrule
    \end{tabular}
    }
    \vspace{-2mm}
    \caption{Statistical significance ($p$-value) and effect size (Cohen's $d$) of the systematic bias in each Big Five trait.}
    \vspace{-2mm}
    \label{table:p_val}
\end{table}
%
%
The non-systematic biases introduced by individual observers will be averaged out in the statistical aggregation process~\cite{simmons2011intuitive,steyvers2014evaluating}.
This phenomenon could be linked to the wisdom of the crowd phenomenon in which aggregating multiple independent judgments often produces estimates superior to individual experts, as random errors tend to cancel each other out when combined \cite{fleenor2006wisdom}. On the other hand, the mean deviation between the observer- and self-report shows the systematic biases that cannot be averaged out in the aggregation process. We note systematic patterns in how observers perceive personality traits compared to self-perceptions (Table~\ref{table:p_val}). Agreeableness, in particular, shows the most significant positive deviation (0.91 point) and the widest spread of ratings, indicating that observers consistently rate individuals as more agreeable than they rate themselves, though with substantial variability in the magnitude of this difference. Conscientiousness also demonstrates moderate positive deviations (0.39), suggesting a consistent tendency for observers to rate these traits slightly higher than self-ratings. In contrast, openness, extraversion, and neuroticism exhibit minor mean deviations with magnitudes no larger than 0.2, indicating that almost no systematic bias exists for these personality dimensions.

\paragraph{Statistical Significance Test}
Further, we conduct a paired-samples t-test to examine the difference between the self- and observer-report ratings of each Big Five personality dimension. We also calculate the Cohen’s $d$ statistic as the standardized effect size, which is the standardized self–observer deviation based on pooled standard deviation \cite{kim2019self}.
Among the five personality dimensions, we identify statistically significant differences for the AGR and CON traits ($p < 0.05$). Specifically, the self-report ratings for agreeableness (AGR) are significantly lower than the observer-report ratings with a large effect size ($d=1.07$), which indicates that the system bias is larger than one full standard deviation. Conscientiousness trait also illustrates a statistically significant deviation between self- and observer-report ratings, but is more moderate in effect size ($d=0.46$).\footnote{For Cohen’s $d$, values around 0.2 indicate a small effect, 0.5 a medium effect, and 0.8 a significant effect \cite{cohen2013statistical}.}
On the other hand, we do not identify significant systematic biases for the other three personality dimensions.

\begin{figure*}[h]
    \centering
    \includegraphics[width=1.01\textwidth]{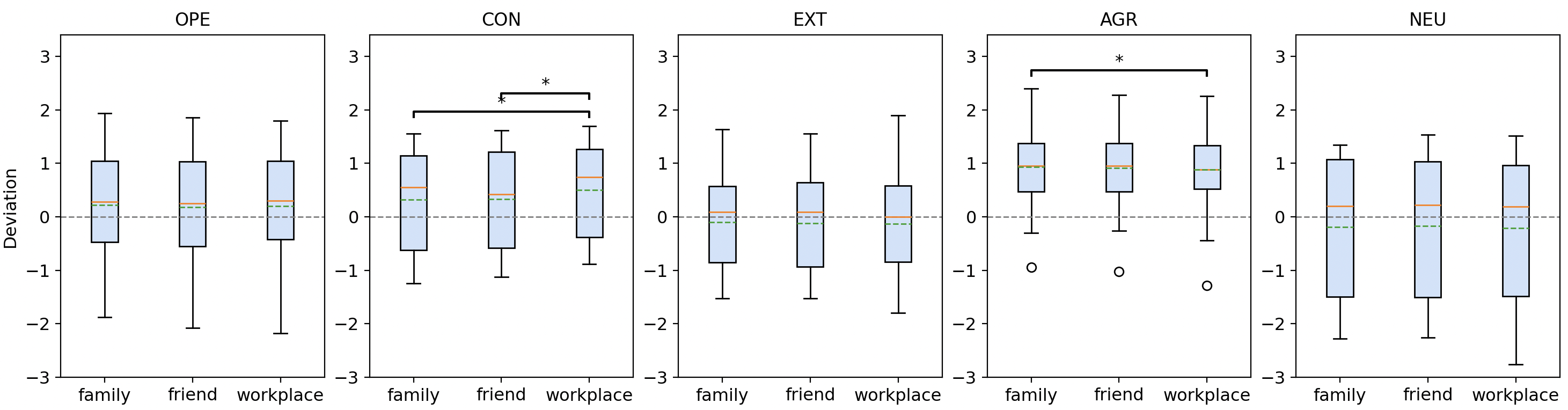}
    \vspace{-6mm}
    \caption{Mean differences between observer and self-reports across Big Five personality traits by relationship context. The orange line represents the median, while the green dotted line shows the mean. Relationships with statistically significant differences ($p-value < 0.05$) are highlighted with asterisks ($*$).} 
    \vspace{-4mm}
    \label{fig:fig1}
\end{figure*}



\paragraph{Comparison to Human Results}
We compare the deviation between self- and observer-reports based on our simulations to previous findings. \citet{kim2019self} conducted a meta-analysis to reveal the mean deviation between human self-report personality and observer-report personality (also called informant-report) of individuals with close relationships. Their work shows that there is only a slight deviation between self- and observer reports, with an effect size smaller than $d<0.27$ across all Big Five dimensions.
This aligns with our results for the personality dimensions of openness, extraversion, and neuroticism. On the other hand, we found that the LLM subject agents possess a significant systematic bias in self-reporting lower agreeableness and conscientiousness scores. Considering that systematic biases exist only in the case of LLM agents but not humans, we speculate that these biases might originate from the alignment training phase. The alignment training encourages LLMs to act according to users' preferences, resulting in an inherent bias in their self-reported personality. The result also suggests that LLM gives self-report personality assessments based on the personality instruction prompts (such as the personality markers used in this experiment), and the scores might not reflect the actual behavior of the agent.

\subsection{Impact of Relationship Type} 
\label{sec:result_relation_type}
\vspace{-1mm}

We now look at the mean deviations between self-report and observer-report ratings under three relationship contexts as illustrated in Figure \ref{fig:fig1}. We observed significant differences between the ratings for conscientiousness and agreeableness across different relationship types. These patterns suggest that relationship context particularly influences the perception of these traits.
Conscientiousness showed significant differences between workplace and friend/family ratings.
Observer agents in a workplace relationship scenario tend to give slightly higher conscientiousness ratings than observers of a family or friend relationship with the subject agent.
Similarly, we observed differences between workplace and family ratings for agreeableness.
Specifically, observer agents in a family relationship tend to give higher agreeableness ratings than observers in a workplace relationship.
On the other hand, other personality dimensions demonstrated consistency across relationship types, with no statistically significant difference in ratings across different relationship types. 





\vspace{-3mm}
\paragraph{Context-Dependency of Personality}
Observers in different relational contexts assign different importance to specific traits. In particular, workplace observers showed distinctly different rating patterns compared to family and friends, particularly for conscientiousness and agreeableness. This divergence likely reflects the multi-faceted and context-dependent nature of personality.
While an individual's personality is generally considered stable, the manifestation of personality may vary across different social contexts \cite{fleeson2015whole}. An extroverted person, for example, may act even more outgoing in front of close friends but adopt a more reserved behavior at work.
Since the observer reports are influenced by the different facets of the person, it is common to observe inconsistency in observer ratings across observers.
Notably, agreeableness and conscientiousness traits are considered the most context-dependent traits among the Big Five dimensions \cite{connelly2010other}. 
Take conscientiousness for example, an individual in a relatively structured situation (e.g., office, meeting, classroom, etc.) may expect it to be easy to concentrate and so may increase his or her level of conscientiousness \cite{fleeson2015whole,nasello2023individual}.

Finally, we found discrepancies in agreeableness and conscientiousness, indicating a higher degree of context dependence of these two traits not only for humans but also for LLM agents. The context-dependent nature of personality highlights the importance of incorporating multiple observer agents in personality assessment. 
By aggregating diverse perspectives of multiple observers, we can construct more comprehensive representations of the agent's personality.

\section{Conclusions}
\label{sec:conclusion}

The study introduced a novel multi-observer framework for personality assessment in LLMs. We verify the validity of our proposed observer-report method, which yields more robust and context-sensitive personality evaluations than traditional self-report methods. 
Further analysis demonstrates the effectiveness of aggregating responses from multiple observer agents, which mitigates individual biases and yields more robust evaluations.
robust and context-sensitive personality evaluations than traditional self-report methods. 
Our experiments reveal that relationship context and observer diversity significantly impact rating patterns, underscoring the importance of tailored personality assessment strategies. Future research will focus on refining the complexity of the scenarios and exploring alternative relationship configurations.


\newpage

\section*{Limitations}

Our simulation framework possesses several limitations that merit discussion. First, the interactive scenarios employed in our study may not fully capture the breadth of personality expressions, particularly for traits like neuroticism or openness that often manifest in less scripted and more intimate contexts. Additionally, the relationship contexts (e.g., family, friends, and workplace) are simplistic compared to the complexity of real-world interpersonal interactions, which could affect the accuracy of our ratings. There are potential discrepancies between the self-reported and observer-reported scales, which complicate the interpretation of our findings. Such findings pave the way for more reliable and nuanced personality assessments in LLMs and support the deployment of psychology-aware agents across diverse social contexts, such as classrooms, relationship counselling \cite{VOWELS2024100078}, mental healthcare \cite{hua2024large}, mental therapy \cite{nie2024llm}, teamwork \cite{arukgoda2023context}, where AI must adapt to individuals’ varying personality traits.


\bibliography{references}

\appendix

\section{Appendix}
\subsection{Agent profile} \label{appendix:agent_profile}
For each subject and observer agent, we randomly pick a name from the $100$ most common names in America\footnote{The list could be found in \url{https://www.ssa.gov/oact/babynames/decades/century.html}}.
The gender feature is assigned accordingly.
For the age feature of the agents, a number is randomly selected from the range of $15$ and $80$.

For subject agents, additional personality instructions are also provided. We follow the setting of~\cite{huang2024rafik} and use the list of $70$ bipolar adjective pairs proposed in the Big Five personality theory along with modifiers like `very', `a bit' to set different levels of personality traits.

In this fashion, we construct the text description of an agent's basic profile as follows:

\begin{tcolorbox}[colback=gray!20!white, colframe=gray!80!black, title=\texttt{Subject Agent Instruction}] 
\vspace{-1mm}
\small
\texttt{Your name is {[}SUBJECT NAME{]}. You are a {[}AGE{]}-year-old {[}GENDER{]}. \\
\\
You have the following personality:\\ {[}PERSONALITY MARKERS{]}.\\ Make sure to reflect your personality traits in your response.}
\end{tcolorbox}

\begin{tcolorbox}[colback=gray!20!white, colframe=gray!80!black, title=\texttt{Observer Agent Instruction}] 
\vspace{-1mm}
\small
\texttt{Your name is {[}OBSERVER NAME{]}. You are a {[}AGE{]}-year-old {[}GENDER{]}.
}
\end{tcolorbox}

\subsection{Prompt Templates} \label{appendix:prompt}
The prompt for the relationship extraction is the following:
\begin{tcolorbox}[colback=gray!20!white, colframe=gray!80!black, title=\texttt{Relation Generation Prompt}] 
\vspace{-1mm}
\small
\texttt{The following are the profiles of two persons X and Y and their relationships:\\
X: {[}SUBJECT BASIC PROFILE{]}\\
Y: {[}OBSERVER BASIC PROFILE{]}\\
\\
Generate {[}\$N{]} diverse {[}RELATION TYPE{]} relations between X and Y. The generated relations must be in the following format: \\
"X and Y are ..."
}
\end{tcolorbox}



The prompt for the scenario extraction is the following:

\begin{tcolorbox}[colback=gray!20!white, colframe=gray!80!black, title=\texttt{Scenario Generation Prompt}] 
\vspace{-1mm}
\small
\texttt{The following are the profiles of two persons X and Y and their relationships:\\
X: {[}SUBJECT BASIC PROFILE{]}\\
Y: {[}OBSERVER BASIC PROFILE{]}\\
relationship: {[}RELATIONSHIP{]}\\
\\
Generate {[}\$K{]} diverse daily life scenarios in which X and Y interact. The scenarios must follow the rules below:\\
1. The scenario should depict a concrete situation where we can observe X's personality.\\
2. DO NOT make presumptions about X's personality in the scenario.\\
3. Generate a short text description of the scenario. For each scenario, also provide which of the Big 5 dimensions it assesses.
}
\end{tcolorbox}

During scenario simulation, we adopt the following instruction for the subject agent.

\begin{tcolorbox}[colback=gray!20!white, colframe=gray!80!black, title=\texttt{Simulation instruction}] 
\vspace{-1mm}
\small
\texttt{{[}SUBJECT BASIC PROFILE{]}\\
{[}LATENT PERSONALITY PROFILE{]}\\
\\
You and {[}OBSERVER NAME{]} (the user) are {[}RELATIONSHIP{]}.\\
Your task is to have a conversation with {[}OBSERVER NAME{]} based on the following scenario:{[}SCENARIO DESCRIPTION{]}\
}
\end{tcolorbox}

The observer agent's instruction is similar but without the personality profile.

Finally, the prompt for self-report personality assessment and observer-report assessment is as follows:

\begin{tcolorbox}[colback=gray!20!white, colframe=gray!80!black, title=\texttt{Prompt for self-report}] 
    \vspace{-1mm}
    \small
    \texttt{{[}SUBJECT BASIC PROFILE{]}\\
    {[}LATENT PERSONALITY PROFILE{]}\\
    \\
    Evaluate the following statement:\\
    {[}STATEMENT{]}.\\
    Rate how accurately this describes you on a scale from 1 to 5 (where 1 = "very inaccurate", 2 = "moderately inaccurate", 3 = "neither accurate nor inaccurate", 4 = "moderately accurate", and 5 = "very accurate"). Please answer using EXACTLY one of the following:  1, 2, 3, 4, or 5.
}
\end{tcolorbox}

\begin{tcolorbox}[colback=gray!20!white, colframe=gray!80!black, title=\texttt{Prompt for observer report}] 
    \vspace{-1mm}
    \small
    \texttt{{[}OBSERVER BASIC PROFILE{]}\\
    The following are some dialogues between you and {[}SUBJECT NAME{]}: \\
    {[}DIALOGUES{]}\\
    \\
    Evaluate the following statement:\\
    {[}STATEMENT{]}.\\
    \\
    Rate how accurately this describes {[}SUBJECT NAME{]} on a scale from 1 to 5 (where 1 = "very inaccurate", 2 = "moderately inaccurate", 3 = "neither accurate nor inaccurate", 4 = "moderately accurate", and 5 = "very accurate"). Please answer using EXACTLY one of the following:  1, 2, 3, 4, or 5.
    }
\end{tcolorbox}

\subsection{Variations of LLM Models and Prompt Formats} \label{appendix:llama}

We also test our observer-report personality assessment framework on different LLM models and prompt variations.

\paragraph{Model Variations}
For model variations, we consider the two other open-sourced models in addition to the GPT-4o model used in the main text of this paper.
\begin{itemize}
    \item \textbf{Qwen2.5}: We adopt the \textit{Qwen/Qwen2.5-72B-Instruct} model developed by Alibaba Cloud~\cite{qwen2.5}.
    \item \textbf{Llama-3}: We adopt the \textit{meta-llama/Meta-Llama-3-70B-Instruct} model developed by Meta~\cite{llama3modelcard}.
\end{itemize}

\begin{figure*}[t]
    \centering
    \includegraphics[width=0.97\textwidth]{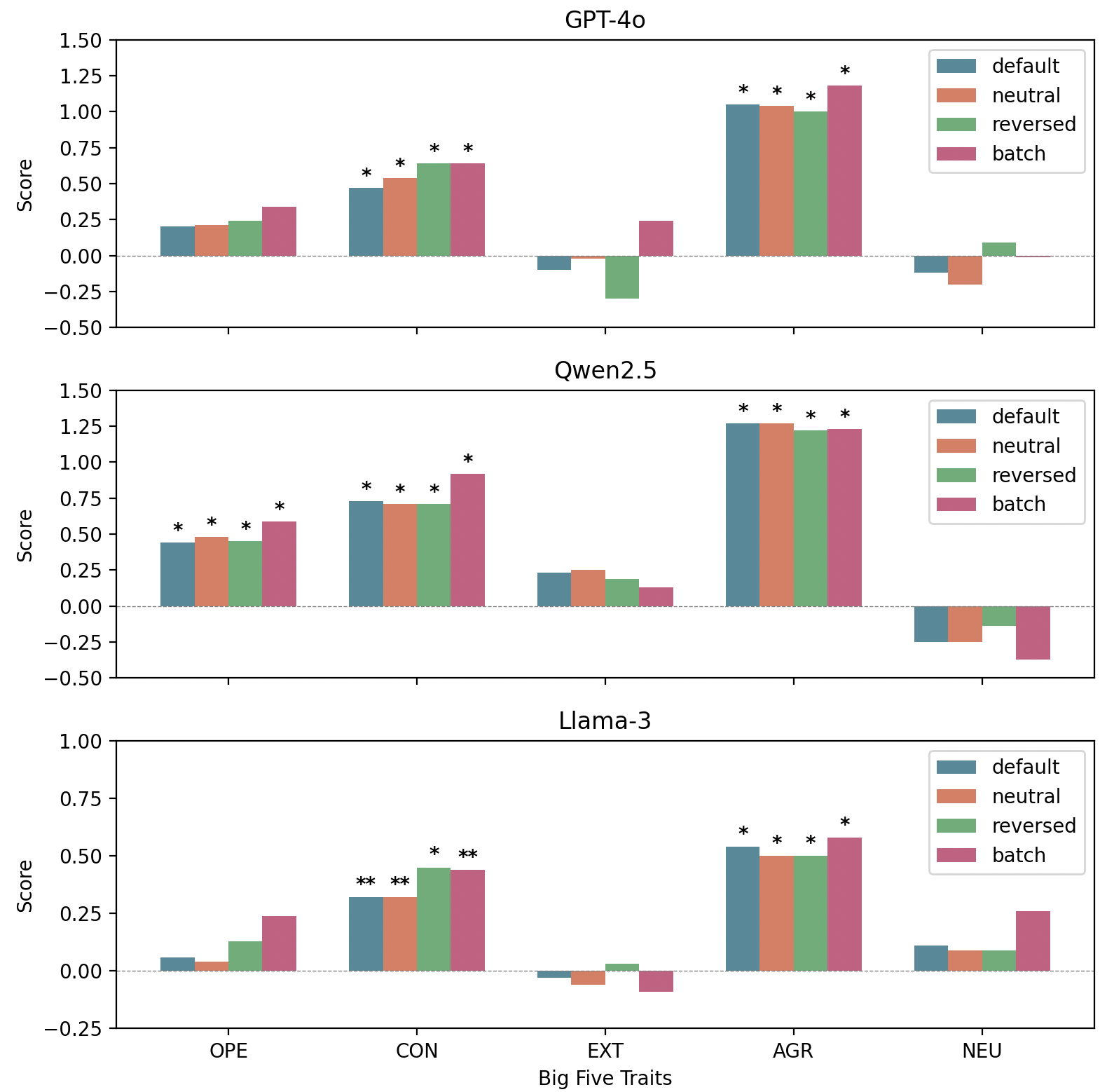}
    \caption{Difference of observer-report and self-report in each Big Five personality dimension for different models and prompt variations. 
    Asterisks indicate differences that are statistically significant (*: p < 0.05, **: p< 0.1).
    } 
    \label{fig:sensitivity}
\end{figure*}

\paragraph{Prompt variations}
Previous works on LLM personality assessment reveal that self-report ratings are highly sensitive to variations in prompt phrasing.
Here, we conduct a sensitivity analysis to observe whether the system biases between self- and observer-reports persist.
We consider the following type of prompt variations.
\begin{itemize}
    \item \textbf{default}: the default prompt setting introduced in Appendix~\ref{appendix:agent_profile} and \ref{appendix:prompt}.
    \item \textbf{neutral}: We introduce variation in the prompt of the subject agent instruction and the observer agent instruction. Specifically, we convert the persona-based style instruction to a more neutral tone as follows.

    \begin{tcolorbox}[colback=gray!20!white, colframe=gray!80!black, title=\texttt{Subject Agent Instruction}] 
        \vspace{-1mm}
        \small
        \texttt{Imagine you are a {[}AGE{]}-year-old {[}GENDER{]} named {[}SUBJECT NAME{]} who have the following personality: \\
        {[}PERSONALITY MARKERS{]}.\\ Make sure to reflect your personality traits in your response.}
    \end{tcolorbox}

    \begin{tcolorbox}[colback=gray!20!white, colframe=gray!80!black, title=\texttt{Observer Agent Instruction}] 
        \vspace{-1mm}
        \small
        \texttt{Imagine you are a {[}AGE{]}-year-old {[}GENDER{]} named {[}SUBJECT NAME{]}}.
    \end{tcolorbox}

\begin{figure*}[t]
  \centering
  \subfigure[Dialogue between the subject and observer agents.]{
    \fbox{\includegraphics[width=0.45\textwidth]{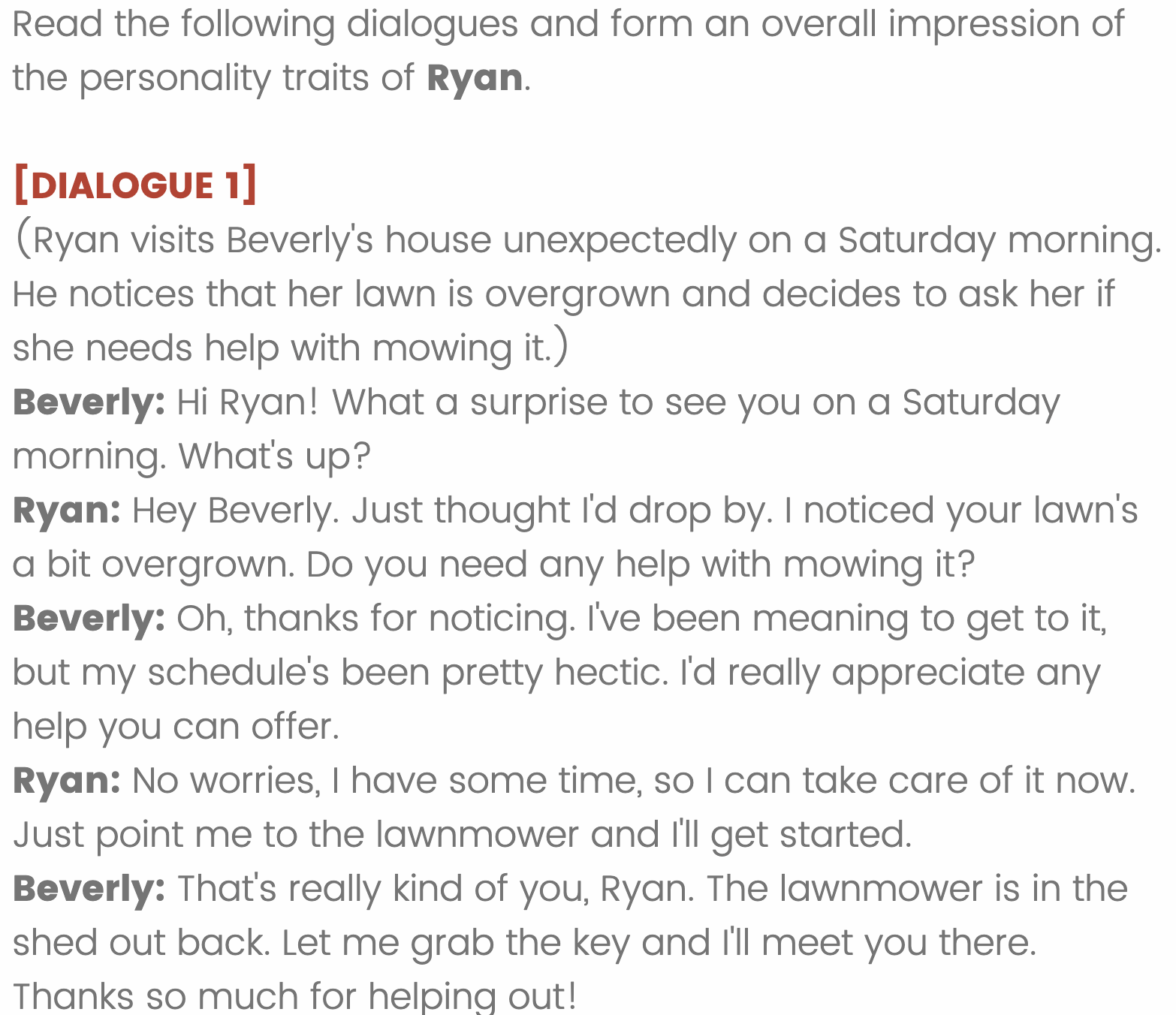}}
    \label{fig:survey_dialog}
  }
  \hfill
  \subfigure[Personality questionnaire issued to human participants.]{
    \fbox{\includegraphics[width=0.45\textwidth]{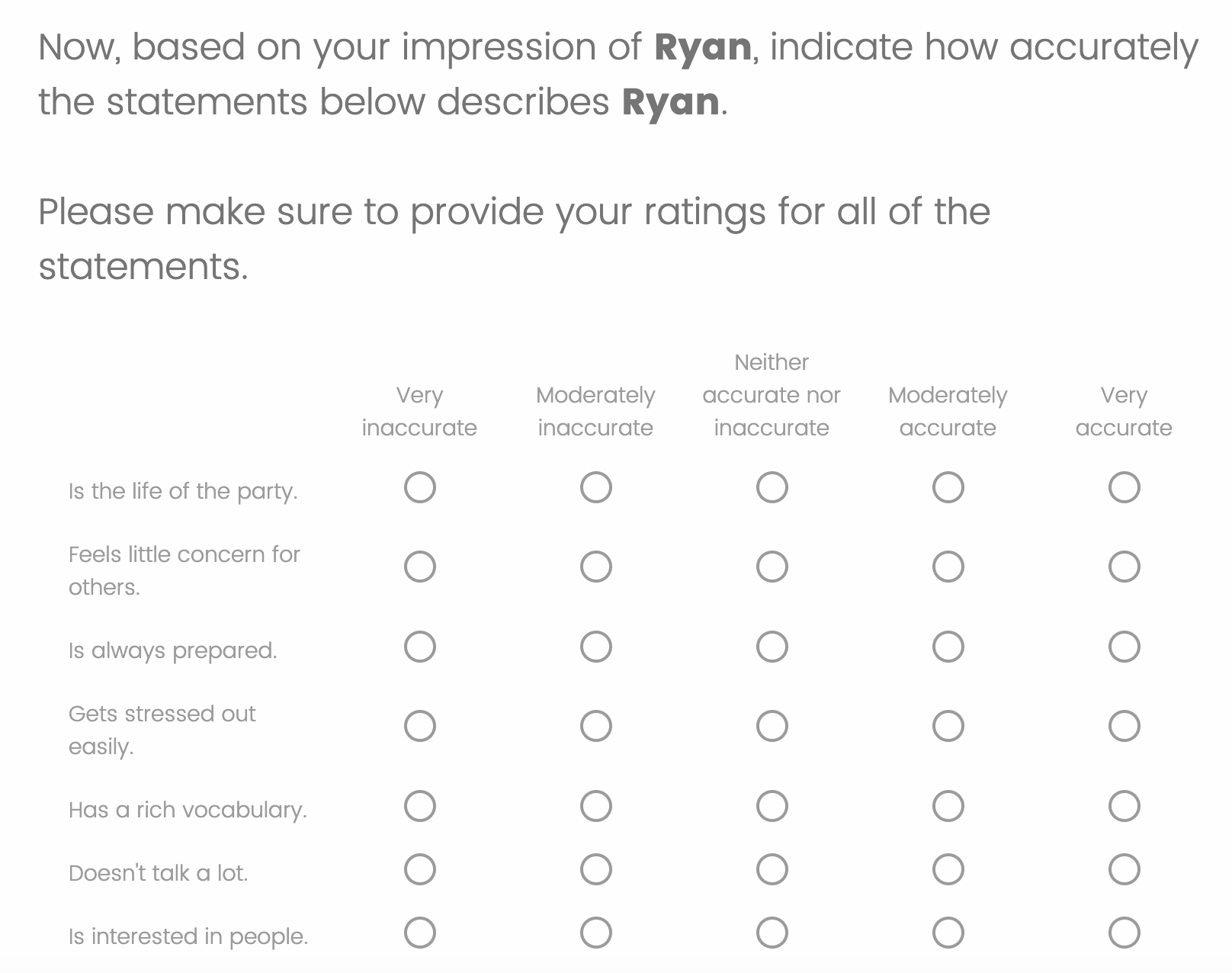}}
    \label{fig:survey_ipip}
  }
  \caption{Screenshots of the example dialogues and the personality questionnaire issued to survey participants.}
  \label{fig:screenshot}
\end{figure*}

    \item \textbf{reversed}: Previous work has also found that LLM responses can be sensitive to the ordering of multiple-choice options. Thus, we try to reverse the order of the 1-5 Likert scale to observe the option-order sensitivity of the deviation between self- and observer-report ratings. Specifically, we reverse the order of the Likert options in the prompt for self- and observer reports:
    \begin{tcolorbox}[colback=gray!20!white, colframe=gray!80!black, title=\texttt{Prompt for self-report}] 
        \vspace{-1mm}
        \small
        \texttt{...
        \\
        Rate how accurately this describes you on a scale from 1 to 5 (where 1 = "very accurate", 2 = "moderately accurate", 3 = "neither accurate nor inaccurate", 4 = "moderately inaccurate", and 5 = "very inaccurate"). ...
        }
    \end{tcolorbox}
    
    \begin{tcolorbox}[colback=gray!20!white, colframe=gray!80!black, title=\texttt{Prompt for observer report}] 
        \vspace{-1mm}
        \small
        \texttt{...
        Rate how accurately this describes {[}SUBJECT NAME{]} on a scale from 1 to 5 (where 1 = "very accurate", 2 = "moderately accurate", 3 = "neither accurate nor inaccurate", 4 = "moderately inaccurate", and 5 = "very inaccurate"). ...
        }
    \end{tcolorbox}
    
    \item \textbf{batch}: We present all 50 items in the personality test questionnaire at once instead of one at a time (as in the default setting). This variable better simulates the real-world personality test settings of human participants and can test the influence of cross-item interference on our results.
\end{itemize}

In Figure \ref{fig:sensitivity}, we report the deviation of observer-report and self-report of different model types and prompt variations.

\begin{table*}[t]
\centering
\resizebox{1.0\linewidth}{!}{%
\begin{tabular}{l*{5}{cc}}
    \toprule
    & \multicolumn{2}{c}{\textbf{OPE}} 
    & \multicolumn{2}{c}{\textbf{CON}} 
    & \multicolumn{2}{c}{\textbf{EXT}} 
    & \multicolumn{2}{c}{\textbf{AGR}} 
    & \multicolumn{2}{c}{\textbf{NEU}} \\
    & self & observer & self & observer & self & observer & self & observer & self & observer \\
    \midrule
    GPT-4o  & 1.275 & \textbf{0.675} & 1.440 & \textbf{0.835} & 0.910 & \textbf{0.698} & 1.185 & \textbf{0.473} & 1.065 & \textbf{0.377} \\
    Qwen2.5 & 1.075 & \textbf{0.363} & 1.290 & \textbf{0.398} & 0.710 & \textbf{0.423} & 1.085 & \textbf{0.385} & 0.815 & \textbf{0.490} \\
    Llama-3 & 1.262 & \textbf{0.850} & 1.273 & \textbf{0.885} & 0.915 & \textbf{0.723} & 1.198 & \textbf{0.690} & 1.002 & \textbf{0.423} \\
    \bottomrule
\end{tabular}
}
\caption{Absolute difference between human ratings and self-ratings (self), and the absolute difference between human ratings and observer-ratings (observer), across different model types.}
\label{table:error_compare_all}
\end{table*}

\begin{table*}[t]
\centering
\resizebox{1.0\linewidth}{!}{%
\begin{tabular}{l*{5}{cc}}
    \toprule 
    & \multicolumn{2}{c}{\textbf{OPE}} 
    & \multicolumn{2}{c}{\textbf{CON}} 
    & \multicolumn{2}{c}{\textbf{EXT}} 
    & \multicolumn{2}{c}{\textbf{AGR}} 
    & \multicolumn{2}{c}{\textbf{NEU}} \\
    & self & observer & self & observer & self & observer & self & observer & self & observer \\
    \midrule
    GPT-4o  & -0.25 & \textbf{0.48} & \textbf{0.47} & 0.43 & \textbf{0.79} & 0.76 & 0.63 & \textbf{0.85} & 0.22 & \textbf{0.42} \\
    Qwen2.5 & -0.16 & \textbf{0.55} & 0.32 & \textbf{0.61} & 0.76 & \textbf{0.90} & 0.52 & \textbf{0.80} & 0.15 & \textbf{0.25} \\
    Llama-3 & -0.23 & \textbf{0.44} & 0.41 & \textbf{0.45} & \textbf{0.78} & 0.60 & 0.58 & \textbf{0.78} & 0.05 & \textbf{0.59} \\
    \bottomrule
\end{tabular}
}
\caption{Spearman’s Rank Correlation coefficients between human ratings and self-ratings (self), and the absolute difference between human ratings and observer-ratings (observer), across different model types.}
\label{table:corr_compare_all}
\end{table*}

Across model types, we observe a similar general tendency in report deviation.
All models show a significant systematic bias in agreeableness, and a slightly moderate level of bias in conscientiousness.
This indicates that the systematic biases in these personality dimensions are universal across model types.
However, there are still some differences in deviation patterns among models.
For instance, we also observe a statistically significant bias in openness ratings for Qwen2.5, which is not observed in other models.
Also, we found that the magnitude of deviation of Llama-3 is smaller compared to other models.
We speculate that the difference in the training process might have given rise to the slight difference.

Across different variations of prompts, we found that using different prompt templates does not have a big impact on the deviation pattern. 

\subsection{Computation Environments and Budget}
For the two open-sourced models, the experiments were conducted on a local server equipped with 4 NVIDIA A100 GPUs (80 GB PCIe) cards. 
Also, we use the VLLM package to accelerate inference with tensor parallelism across 4 GPUs.
We used mixed-precision (float16) inference.

For GPT-4o, the simulation process to collect the self-report and observer-reports for one subject agent costs around 2.9\$.

For all models, we set the temperature to 1.0 during the simulation process. When answering personality questionnaires, the temperature is set to 0.0.

\subsection{Human Ratings} \label{sec:human}

\paragraph{Data collection via crowd-sourcing}
We collected human ratings via crowd-sourcing platform Qualtrics\footnote{\url{https://www.qualtrics.com/}} and Prolific\footnote{\url{https://www.prolific.com/}}.
Each consenting participant was presented with five dialogues between a subject and observer LLM agents (See Figure~\ref{fig:survey_dialog}). After reading the dialogues, the participant is asked to rate the designated subject agent's personality by answering the 50-item IPIP questionnaire used in our LLM-based experiments (Figure~\ref{fig:survey_ipip}). We collected a total of 16 valid data samples.

We recruited 16 native English speakers residing in the United Kingdom, the United States, New Zealand, Canada, and Australia. To ensure linguistic proficiency, only individuals who self-identified as native English speakers in the aforementioned countries were eligible. The average completion time was approximately 15 minutes, and each participant received GBP 2.25 upon complete submission.

In addition to detecting biases in self-reports, this work could detect biases across LLMs. We hypothesize that due to their different alignment training strategies, they exhibit different tendencies regarding trait clusters such as (AGR, OPE, CON), EXT, and NEU.


\paragraph{Evaluation based on human ratings}
Based on the collected human ratings data, we calculate the absolute difference between human and self- and observer-report ratings, respectively (Table~\ref{table:error_compare_all}).
We can see that for all model types and personality dimensions, the observer-report ratings show a smaller discrepancy compared to self-report ratings.

We also calculate the agreement between human and self- and observer-report ratings, based on Spearman's rank correlation coefficient (Table~\ref{table:corr_compare_all}).
For the personality dimensions of openness, agreeableness, and neuroticism, we observe a higher agreement between human and observer ratings.
For conscientiousness, we observe a higher agreement for self-report, but the difference is marginal.
For extraversion, self-report ratings result in a higher agreement in the case of GPT-4o (marginal difference) and Llama-3 (0.18 points higher).

Based on the results above, we can conclude that observer report ratings align better with human evaluation in general.

\begin{table*}[t]
    \centering
    \small
    \begin{tabular}{p{2.3cm}p{13cm}}
    \toprule
    \textbf{Subject Profile} & \texttt{\{name: Ethan, age: 29, gender: male\}}\\
    \midrule
    \textbf{Observer Profile} & \texttt{\{name: Jacob, age: 52, gender: male\}} \\
    \midrule
    \textbf{Relation Context} &  Workplace\\
    \midrule
    \textbf{Relation} & Target is the Observer’s mentee. \\
    \midrule
    \textbf{Scenario} & Jacob is faced with a difficult decision regarding project resources and seeks Ethan's opinion. He emphasizes the importance of considering the team's wellbeing and long-term goals while making the decision. \\
    \midrule
    \textbf{Dialogue} & 
        Observer: Hey Ethan, I hope you're doing well. I've been faced with a tough decision lately regarding the allocation of resources on our current project. It's a bit of a balancing act between meeting short-term demands and ensuring the well-being of our team while keeping an eye on our long-term goals. I'd really appreciate your opinion on this. What do you think is the best way to approach this situation?
        \vspace{1mm}

        \textbf{Subject: Well, Jacob, I'm not really great at this kind of thing, but maybe just stick to what's already being done. Trying something new or adventurous isn't usually worth the risk, and people might not be happy with changes. Maybe just distribute resources evenly and hope for the best. Sorry if that's not much help.}
        \vspace{1mm}

        Observer: No need to apologize, Ethan. Your perspective is valuable, and I appreciate your honesty. Sticking to what's tried and tested can indeed be a safe approach to maintain stability. Perhaps a small adjustment could be to actively check in with the team to ensure they feel supported and communicate clearly about any changes. It might help to catch potential issues early. Do you think that could work, or do you have any other thoughts?
        \vspace{1mm}
        
        \textbf{Subject: Yeah, that might work. Checking in sounds reasonable, but as long as it's not too complicated. Simple communication could help, I guess. Just make sure it doesn't get too involved or overwhelming. Whatever keeps things stable is probably the safest bet.}
        \vspace{1mm}

        Observer: I completely agree with you, Ethan. Keeping communication straightforward and not overwhelming is key. I'll make sure to keep our check-ins simple to maintain stability without adding undue pressure. Thanks for your input. It's always great to have your perspective.
        \\
    \bottomrule
    \end{tabular}
    \caption{Case study: mentor–mentee dialogue on project resource allocation.}
    \label{table:case}
\end{table*}

\begin{table*}[t]
\centering
\resizebox{0.7\linewidth}{!}{%
\begin{tabular}{lccccc}
\toprule
 & \textbf{OPE} & \textbf{CON} & \textbf{EXT} & \textbf{AGR} & \textbf{NEU} \\
\midrule
\textbf{Subject Latent Personality Profile} & 1 & 4 & 2 & 1 & 2 \\
\midrule
\textbf{Subject Self-report} & 1.0 & 2.6 & 1.5 & 1.7 & 2.0 \\
\textbf{Individual Observer-report} & 1.8 & 2.8 & 1.3 & 2.8 & 1.5 \\
\textbf{Aggregated Observer-report} & 1.8 & 2.6 & 1.6 & 3.0 & 2.4 \\
\bottomrule
\end{tabular}
}
\caption{Latent personality level, self-report and observer-report scores for all Big Five dimensions corresponding to the example in Table~\ref{table:case}. }
\label{table:case-scores}
\end{table*}

\subsection{Case Study}

In this section, we conducted a case study of our proposed method. Table~\ref{table:case} presents an example of the simulation dialogue between a pair of subject and observer agents, \textit{Ethan} and \textit{Jacob}, respectively. Starting from the automatically generated agent profiles and the relationship between them, scenarios are generated automatically. Then, we conduct a dialogue simulation of the interaction between the agents under these scenarios. For space limitations, we only present one example of the scenarios and its corresponding simulated dialogue. After the scenario simulation process, the observer agent is asked to give ratings on the subject agent's personality.

Table~\ref{table:case-scores} shows the latent personality level given as instructions to the subject agent and the self- and observer-report scores. Specifically, the \textbf{Individual Observer-report} scores are the ratings the observer Jacob offers, and the \textbf{Aggregated Observer-report} scores are the average scores of all observers' ratings.

Table~\ref{table:case} and~\ref{table:case-scores} present an intriguing case where an agent is assigned a “very disagreeable” personality (with the lowest AGR level of 1).
We can see that the self-report score given by the subject agent itself gives a low agreeableness score ($1.7$), where the observer agents give a moderate agreeableness score ($3.0$). However, by observing the simulated dialogue in Table~\ref{table:case}, we can see that the real behavior manifested by the subject agent \textit{Ethan} does not reflect such extreme disagreeableness quality suggested by the self-report score.
This discrepancy underscores the importance of examining behaviors beyond self-report questionnaires.

\end{document}